\documentclass[1p]{elsarticle}

\usepackage{amsthm}
\usepackage{amssymb}
\usepackage{amsmath}
\usepackage{amsfonts}
\usepackage{wrapfig}

\usepackage{cite}
\usepackage{float}
\usepackage{graphicx}
\usepackage{mwe}
\usepackage{graphbox}
\usepackage{setspace}
\usepackage{epstopdf}
\usepackage{epsf}
\usepackage{enumitem}
\usepackage{hyperref}

\usepackage{bm}
\usepackage{acronym}

\usepackage[ruled,vlined]{algorithm2e}

\theoremstyle{remark}	\newtheorem{theorem}{Theorem}
\theoremstyle{remark}	
\theoremstyle{remark}	

\biboptions{sort&compress}

\begin{document}

\hspace{-15pt}{\large Less is More: Rethinking Few-Shot Learning and Recurrent Neural Nets} \\
Deborah Pereg, Martin Villiger, Brett Bouma, Polina Golland 
\vspace{1cm}

\title{Information Theoretic Perspective on Sample Complexity} %

\author{Deborah Pereg
\footnote{D. Pereg was with the Wellman Center for Photomedicine MGH, Harvard Medical School, and MIT CSAIL (e-mail: dvorapereg@gmail.com).}
}

\begin{abstract}
The statistical supervised learning framework assumes an input-output set with a joint probability distribution that is reliably represented by the training dataset. The learning system is then required to output a prediction rule learned from the training dataset's input-output pairs. In this work, we investigate the relationship between the sample complexity, the empirical risk and the generalization error based on the asymptotic equipartition property (AEP) \citep{Shannon:1948}. We provide theoretical guarantees for reliable learning under the information-theoretic AEP, with respect to the generalization error and the sample size in different settings.

\vspace{5mm}
\end{abstract}

\maketitle

\section{Introduction}

Deep neural networks (DNNs) have led to state-of-the-art results, spanning through numerous fields of knowledge. Nevertheless, a clear theoretical understanding of important aspects of artificial intelligence (AI) is still missing. Furthermore, there are many challenges concerning the deployment and implementation of AI algorithms in practical applications, primarily due to high computational complexity and insufficient generalization. Concerns have also been raised regarding the carbon footprint of training large scale deep learning systems \citep{Strubell:2020}. Improving sample efficiency and generalization have been the center of attention and efforts of many in the industrial and academic research community \citep{BengioLecun:2007,Vincent:2008,Bengio:2009,Zhang:2021}. %

Typically, in statistical learning \citep{BenDavid:2014,Vapnik:1999}, it is assumed that the instances of the training data are generated by some probability mass function. 
For example, we can assume a training input set $\Psi_{Y}=\{\{\mathbf{y}_{i}\}_{i=1}^m : \mathbf{y}_{i} \sim P_Y\}$, such that there is some correct target output $\mathbf{x}$, unknown to the learner, and each pair $(\mathbf{y}_i,\mathbf{x}_i)$
in the training data $\Psi$ is generated by first sampling a point $\mathbf{y}_i$ according
to $P_Y(\cdot)$ and then labeling it. The examples in the training set are randomly chosen and, hence, independently
and identically distributed (i.i.d.) according to the distribution $P_Y(\cdot)$.
We have access to the training error (also referred to as empirical risk), which we normally try to minimize. The known phenomenon of overfitting is when the learning system fits perfectly to the training set and fails to generalize. In classification problems, probably approximately correct (PAC) learning defines the minimal size of a training set required to guarantee a PAC solution. The sample complexity depends on the accuracy of the labels and a confidence parameter. It is also a function of properties of the hypothesis class. To describe generalization we normally differentiate between the empirical risk (training error) and the true risk. 

Recent works \citep{Tishby:2015,shwartz:2017,shwartz:2018} introduced the information bottleneck (IB) principle in the context of supervised learning  and demonstrated that the convergence of DNNs' layers
follows the IB optimal bound.
Shwartz et al. (2018) \citep{shwartz:2018} show that for a high dimensional input and typical input sequences, the mutual information up to representation level (layer) $T$ controls the complexity of the problem, given a generalization error. 

In this work we investigate theoretical aspects of sample complexity, based on the information-theoretic asymptotic equipartition property (AEP) \citep{ThomasCover:2006}. We show that there exists a relatively small set that can empirically represent the input-output data distribution for learning. Our work can hopefully illuminate and pave a path towards empirical possibilities of learning with limited ground truth training data. 

\section{Background}\label{sec2}

Hereafter, we use the notation $x^n$ to denote a sequence $x_1,x_2,...,x_n$, and $\mathbf{x}\in\mathcal{X}^{n \times 1}$ to denote a vector with $n$ entries. As known, in information theory, a stationary stochastic process $u^n$ taking values in some finite alphabet $\mathcal{U}$ is called a source. In communication theory we often refer to discrete memoryless sources (DMS) \citep{Kramer:2008,ThomasCover:2006}. However, many signals, such as image patches, are usually modeled as entities belonging to some probability mass function forming statistical dependencies (e.g., a Markov random field (MRF) \citep{Roth:2009,Weiss:2007}) describing the relations between data points in close spatial or temporal proximity.
Here, we will briefly summarize the AEP for ergodic sources with memory \citep{Austin:2017}. Although the formal definition of ergodic process is somewhat complex, the general idea is simple. ``In an ergodic process, every sequence that is produced by the process is the same in statistical properties" \citep{Shannon:1948}. The symbol frequencies obtained from particular sequences generated by the process, will approach a definite statistical limit, as the lengths of the sequences is increased.
More formally, we assume an ergodic source with memory that emits $n$ symbols from a discrete and finite alphabet $\mathcal{U}$, with a probability mass function $P_U(u_1,u_2,...,u_n)$.
We recall a theorem \citep{Breiman:1957}, here without proof.
\begin{theorem}[Entropy and Ergodic Theory \citep{Breiman:1957}] %
\label{Theorem 1.1} Let $(u_n)_{n \in \mathbb{Z}}$ be a stationary ergodic process ranging over a finite alphabet $\mathcal{U}$, then there is a constant $H$, defining the entropy rate of the source,
\begin{equation*} 
H = \lim_{n\rightarrow\infty} -\frac{1}{n} \log_2 P_U(u_1,...,u_n).
\end{equation*} \end{theorem}

Intuitively, when we observe a source with memory over several time units, the uncertainty grows more slowly as $n$ grows, because once we know the previous source's entries, the dependencies reduce the overall conditional uncertainty. 
The entropy rate $H$, which represents the average uncertainty per time unit, converges over time. This, of course, makes sense, as it is known that {\small $H(X,Y)\leq H(X)+H(Y)$}. In other words, the uncertainty of a joint event is less than or equal to the sum of the individual uncertainties.
The generalization of the AEP to arbitrary ergodic sources is as following \citep{Breiman:1953}.
\begin{theorem}[Shannon McMillan (AEP)\citep{Breiman:1953}]
\label{Theorem 1.2}
For $\epsilon>0$, the typical set $A^n_\epsilon$ with respect to the ergodic process $P_U(u)$ is the set of sequences $\mathbf{u}=(u_1,u_2,...,u_n)\in \mathcal{U}^n$ obeying
\begin{enumerate}
\item $\lim_{n \rightarrow \infty} \mathrm{Pr}[\mathbf{u} \in  A^n_\epsilon]=1$.
\item $2^{-n(H+\epsilon)} \leq P_U(\mathbf{u}) \leq 2^{-n(H-\epsilon)}$.
\item $(1-\epsilon)2^{n(H-\epsilon)} \leq |A^n_\epsilon| \leq 2^{n(H+\epsilon)}$, for $n$ sufficiently large.
\end{enumerate}
$|A|$ denotes the number of elements in the set $A$, and $\mathrm{Pr}[\mathcal{A}]$ denotes the probability of the event $\mathcal{A}$ .
\end{theorem}
In other words, if we draw a random sequence $(u_1,u_2,...,u_n)$, the typical set occurs approximately with probability 1. All elements of the typical set $A^n_\epsilon$ are approximately equally probable, and the number of elements of the typical set is approximately {\small $2^{nH}$}. This property is called the asymptotic equipartition property (AEP). In information theory the AEP is considered as the analog of the law of large numbers \citep{ThomasCover:2006}. %
The notion of a typical sequence was first introduced in 1948 by Shannon in his paper ``A Mathematical Theory of Communication'' \citep{Shannon:1948}.
Intuitively, the typical sequences $u^n$ are the sequences whose
\textit{empirical} probability distribution is close to $P_U(\cdot)$.

As mentioned, the entropy rate is more often used for discrete memoryless sources (DMS), yet
every ergodic source has the AEP \citep{Breiman:1953}. Entropy typicality applies also to continuous random variables with a density $p_U$ replacing the discrete probability $P_U(u^n)$ with the density value $p_U(u^n)$. 
The AEP leads to Shannon's source coding theorem stating that the average number of bits required to specify a symbol in a sequence of length $n$, when we consider only the most probable sequences, is $H$.
And it is the foundation for the known rate-distortion theory and channel capacity. 
 
The AEP property divides the space of all possible sequences into two sets: the
typical set, where the sample empirical entropy is close to the true entropy, and the
non-typical set of all other sequences. Furthermore, the average behavior of any large sample is determined by the typical sequences properties, because 
any property that is true for the typical sequences will be then true with high probability \citep{ThomasCover:2006}. 
Consequently, we will show that under certain assumptions, the typical set controls the balance between the empirical risk, the generalization error and the sample complexity, in a general learning framework. 
This result establishes a fundamental resource characterization of the supervised learning framework.

\section{AEP Perspective on Sample Complexity and Generalization Error}\label{sec3}

\begin{figure}
    \centering
    \includegraphics[width=0.7\textwidth]{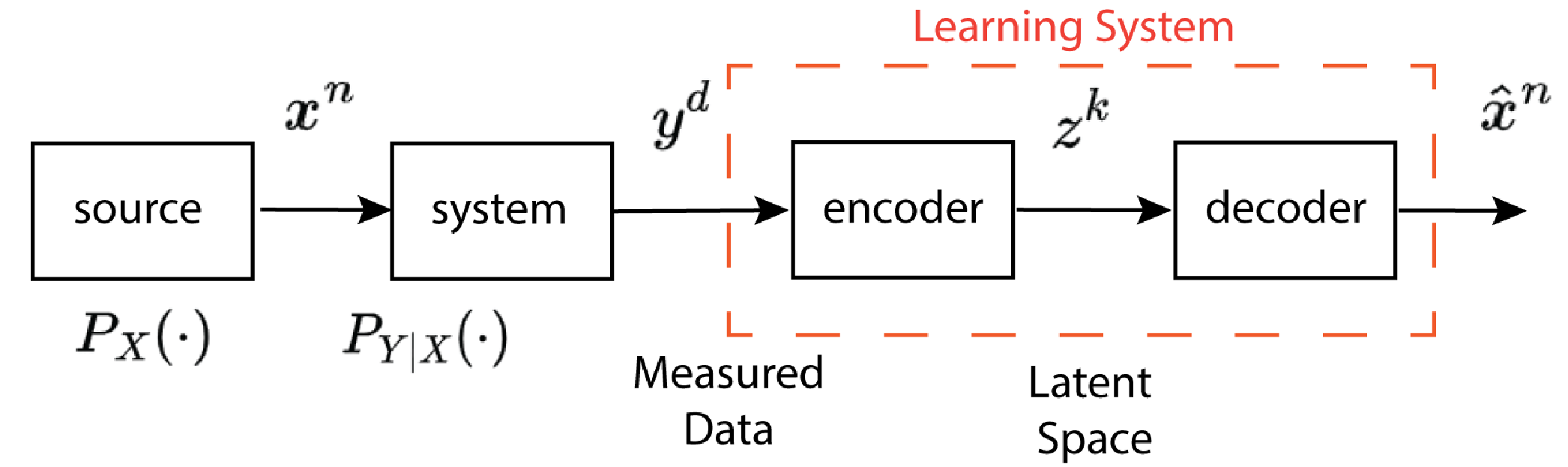}
		\caption{\footnotesize The learning system problem.}
		\label{fig1}
\end{figure}

Let us assume a training set $\Psi=\{\mathbf{y}_{i},\mathbf{x}_{i}\}_{i=1}^m$, where $\mathbf{x}_i \in \mathcal{X}^{n\times 1}$ are sampled from $A^n_\epsilon(P_X)$ and paired with $\mathbf{y}_i \in \mathcal{Y}^{d\times 1}$ by some function as ground truth. The learning system is trained to output a prediction rule $\mathcal{F}: \mathcal{Y}^d \rightarrow \mathcal{X}^n$. Assume an algorithm 
that trains the predictor by minimizing the training error (empirical error or empirical risk). %
Assuming a discrete source $P_X(\cdot)$ that emits i.i.d sequences $x^n$ of symbols (for example: patches in an image, segments of an audio signal, etc.), the estimator has access only to the observed signal $y^d$. 
In an inverse problem $y^d$ would be a degraded signal originating in $x^n$, where the relationship between $x^n$ and $y^d$ could be linear or non-linear, with or without additive noise, such that generally $y^d=g(x^n)+e(x^n)$, where $g(\cdot)$ and $e(\cdot)$ are functions of $x^n$.
Given $y^d$, we produce an estimate $\hat{x}^n$. 
However, the following proofs are not restricted to this framework. 
A possible problem setting with an encoder-decoder learning system is illustrated in Fig.~\ref{fig1}.
For the sake of the following theoretical analysis we restrict the mapping $\mathcal{F}: \mathcal{Y}^d \rightarrow \mathcal{X}^n$ to be a surjective function. Namely, for every $x^n$, there is a $y^d$ such that $f(y^d) = x^n$. In other words, every element $x^n$ is the image of at least one element of $y^d$. It is not required that $y^d$ be unique.
In the presence of noise this condition can only be met if the noise's power is under a certain threshold.
The goal is to prove that learning with $A^n_\epsilon$ is sufficient for generalization over the entire distribution with the same generalization error.
Denote a sample size $|\Psi|=m$ that is required to train a predictor $\mathcal{F}_{\Psi}: \mathcal{Y}^d \rightarrow \mathcal{X}^n$, such that $\Psi \sim P_{X,Y}$. 
An algorithm minimizes the \textit{training error} (\textit{empirical error} or \textit{empirical risk})
\begin{equation}\label{2.1}
\mathcal{L}_{\Psi}(\mathcal{F}_{\Psi}) = \frac{1}{m} \sum_{i=1}^m \ell(\mathcal{F}_{\Psi}(\mathbf{y}_i),\mathbf{x}_i),
\end{equation}
where $0 \leq \ell(\hat{\mathbf{x}},\mathbf{x}) \leq 1$ is some loss function. 
The empirical error over the training set at the end of the training, for the specific trained predictor $h_{\Psi}$ is $\mathcal{L}_{\Psi}(h_{\Psi})\leq \Delta_m<<1$. The true error, or the \textit{generalization error}, in this setting is
\begin{equation}\label{2.2}
\mathcal{L}(h_{\Psi}) = E_{(x,y) \sim P_{X,Y}} \ell \big(h_{\Psi}(\mathbf{y}),\mathbf{x}\big),
\end{equation}
where $E_{(x,y) \sim P_{X,Y}}(\cdot)$ denotes the expectation over $P_{X,Y}$.
We evaluate the ability of a trained learner to generalize well by the upper bound on the generalization error,
$\mathcal{L}(h_{\Psi})\leq \Delta$. 

\begin{theorem}[AEP learning for systems with a surjective mapping]
\label{Theorem 2}
Assume the generalization error of the trained predictor $\mathcal{F}: \mathcal{Y}^d \rightarrow \mathcal{X}^n$ over the output typical set $A^n_\epsilon(P_X)$ or over the input typical set $A^d_\epsilon(P_Y)$, for $n$ sufficiently large or $d$ sufficiently large, respectively, is at most $\varepsilon^\mathcal{F}_A$.
Then, the generalization error of the trained predictor $\mathcal{F}: \mathcal{Y}^d \rightarrow \mathcal{X}^n$ over the entire distribution is at most $\varepsilon^\mathcal{F}_A$. 
\end{theorem}
\begin{theorem}[sample size for learning with a surjective mapping]
\label{Theorem 3}
Assume a training set that is generated by randomly drawing samples from $P_X$ and labeling them by the target function $g(\cdot)$,  
$\Psi = \big\{ \{\mathbf{y}_{i},\mathbf{x}_{i}\}_{i=1}^m : 
\mathbf{x}_i \sim P_X, \quad \mathbf{y}_i= g(\mathbf{x}_i), \ \mathbf{x}_i \in \mathcal{X}^{n\times 1},\ \mathbf{y}_i \in \mathcal{Y}^{d\times 1} \big\}$. $g(\cdot)$ is a deterministic function, and $f(\cdot)=g^{-1}(\cdot)$ is a surjective function. Assume the predictor was trained successfully to yield $\mathcal{L}_{\Psi}(h_{\Psi})\leq\Delta_m << 1$.
For $d$ sufficiently large, the sample size $m$ and the true error obey
{\small
\begin{flalign}\label{2.3}
\nonumber
&\mathcal{L}(h_{\Psi}) \leq \Delta_m, \quad \mathrm{if} \ m \geq 2^{dH(Y)} \\  
&\mathcal{L}(h_{\Psi})\leq \Delta, \quad ~~\mathrm{if}  \  2^{dH(Y)}\frac{1-\Delta}{1-\Delta_m} \leq m<2^{dH(Y)}
\end{flalign}
}
where $\Delta_m \leq \Delta \leq 1$, and $\Delta$ is the generalization error bound.
\end{theorem} 
As long as the sample size is larger or equal to the typical set size, we are guaranteed to have a generalization error that is as small as the training error. Otherwise, the upper bound on the generalization error is determined by the ratio between the sample size and the input typical set size.
\begin{theorem}[sample size for learning with a surjective mapping in noisy environment]
\label{Theorem 4}
Assume a training set that is generated by randomly drawing samples from $P_X$ and labeling them by the target function such that,  
$\Psi = \big\{ \{\mathbf{y}_{i},\mathbf{x}_{i}\}_{i=1}^m : 
\mathbf{x}_i \sim P_X, \quad \mathbf{y}_i= g(\mathbf{x}_i)+\mathbf{w}_i, \ \mathbf{x}_i \in \mathcal{X}^{n\times 1},\mathbf{w}_i \in \mathcal{W}^{n\times 1},\ \mathbf{y}_i \in \mathcal{Y}^{n\times 1}\big\}$, 
where $g(\cdot)$ is a deterministic known or unknown function, and $\mathbf{w}_i$ is an additive i.i.d noise. 
For $n$ sufficiently large, the sample size $m$ and the true error obey
\begin{flalign}\label{2.4}
\nonumber
&\mathcal{L}(h_{\Psi}) \leq \Delta_m, \quad \mathrm{if} \quad m \geq 2^{nI(X;Y)} \\  
&\mathcal{L}(h_{\Psi})\leq \Delta, \quad ~~\mathrm{if}  \quad 2^{nI(X;Y)}\frac{1-\Delta}{1-\Delta_m} \leq m<2^{nI(X;Y)},
\end{flalign}
where $\Delta_m \leq \Delta \leq 1$, $\Delta$ is the generalization error bound, and $I(X;Y)=H(Y)-H(Y|X)$ is the mutual information between $X$ and $Y$. \\
\end{theorem} 
In this case, if $y^n$ and $x^n$ are jointly typical, then we can resolve an input $y^n$ as $x^n$. There are approximately $2^{nH(Y|X)}$ equally probable $y^n$ sequences, for each typical output sequence $x^n$. We assume
that no two $x^n$ sequences correspond to the same $y^n$ sequence,
(otherwise, the learner will not be able to decide which $x^n$ sequence it originated from).
The total number of possible typical $y^n$ sequences is approximately $2^{nH(Y)}$. This set
is split into sets of size $2^{nH(Y|X)}$,
associated with different $x^n$ sequences. Therefore, the total number of distinguishable sets is less than or equal
to {\small $2^{n(H(Y)-H(Y|X))}=2^{nI(X;Y)}$}.
Hence, we can have at most $2^{nI (X;Y)}$ disjoint sequences of length $n$.

Theorem~\ref{Theorem 3} is a specific case of Theorem~\ref{Theorem 4}, since in our setting, in the absence of noise {\footnotesize $I(X;Y)=H(Y)$}. Note that when given {\small$EY^2<\sigma^2_x+\sigma^2_w$}, where $\sigma^2_x$ and $\sigma^2_w$ are the output variance and the noise variance, respectively, we know {\small $I(X;Y)\leq \frac{1}{2}\log(1+\sigma^2_x / \sigma^2_w)$}. The proofs for Theorems \ref{Theorem 2}-\ref{Theorem 4} can be found in the Appendices. The theorems and their proofs can be generalized to the continuous case with $A^n_\epsilon(p_X)$ and differential entropy $h(p_X)$,
although it is often possible to assume that $\mathcal{X}$ and $\mathcal{Y}$ are discrete alphabets as a result of quantization of real values, and therefore  we can discuss discrete entropy.
We have assumed $0 \leq \ell(\hat{\mathbf{x}},\mathbf{x}) \leq 1$, without loss of generality. When the loss obeys $\big\{0 \leq \ell(\hat{\mathbf{x}},\mathbf{x}) \leq \ell_{\mathrm{max}}
:~\ell_{\mathrm{max}}> 1\big\}$, our results have a similar form, but $\mathcal{L}(h_{\Psi})\leq \Delta$ when $2^{nI(X;Y)}\frac{1-\tilde{\Delta}}{1-\tilde{\Delta}_m} \leq m<2^{nI(X;Y)}$ such that $\tilde{\Delta}=\Delta/\ell_{\mathrm{max}}$,
$\tilde{\Delta}_m=\Delta_m/\ell_{\mathrm{max}}$ and $\tilde{\Delta}_m \leq \tilde{\Delta} \leq 1$.

The size of the typical set is exponential by $n$, and therefore could be relatively large. Consequently, it could be argued that the above results are not necessarily of great practical significance for improving sample efficiency. Nevertheless, it is important to keep in mind that the AEP property is widely employed in many other applications and enables significantly decreased complexity, such as in data compression and source coding. Although the size of the typical set is exponential by $n$, compression algorithms, such as JPEG \citep{Wallace:1991}, are able to compress an image by a factor of 10. In the noiseless case, there are $2^{n\log_2 r}$ possible output sequences, where $r=|\mathcal{Y}|$. The AEP allows us to train with significantly less examples, because we only try to generalize over $2^{n H(Y)}$ possible sequences. 

The IB framework \citep{Tishby:2015,shwartz:2017,shwartz:2018} aims to optimize a representation by maximizing the mutual information between the input to a representation layer, while compressing the input. Intuitively, attempting to extract from the input only information relevant to the target. Our results, provide an alternative point of view for high dimensional data, describing the relationships between the sample complexity, the empirical risk, and the generalization error, depending on the input-output mutual information, yet independent of the underlying representation captured by the learner.

The above results may lead to further information-theoretic interpretations and insights, in relation with channel coding and source coding. The capacity of a channel defines the optimal transmission rate of communication over a channel $p_{Y|X}(y|x)$. Shannon's channel coding theorem \citep{Shannon:1948} states that channel capacity is $C(p_{Y|X})=\max_{p_X}I(X;Y)$. 
In rate-distortion theory, we are trying to compress the source input under a constraint on the distortion. 
The optimal compression rate is $R(D)=\min_{p_{\hat{X}|X}}I(X;\hat{X}) \ \mathrm{s.t} \ Ed(\hat{x},x)\leq D$, where $x$ is the source, $\hat{x}$ is the decoded signal, $d(\hat{x},x)$ is a distortion measure and $D$ is a given distortion. 
Thus, in channel transmission, we wish to ﬁnd the largest set of codewords that have a large
minimum distance between codewords, whereas in rate-distortion, we try
to ﬁnd the smallest set of codewords that covers the entire space. The direct connection with our results is still an open problem, as in our problem setting, given an input-output distribution, we are trying to design a system that captures the mapping between them, by learning from fewer examples as possible (lower sample complexity). We have shown here that the generalization error bound depends on the sample complexity relative to a given input-output mutual information.\\
\textit{Definition} 6 (Learning Rate).
Define $\mathcal{R}>0$ as the learning rate representing the sample complexity relative to the input-output data dimensions $n$,  
such that
$m=2^{n\mathcal{R}}$. \\
Thus, we have established in Theorem 5 that 
\begin{equation*}
\forall \mathcal{R} \geq I(X;Y):\mathcal{L}(h_{\Psi})=E_{(x,y) \sim P_{X,Y}} \ell \big(h_{\Psi}(\mathbf{y}),\mathbf{x}\big)  \leq \Delta_m,
\end{equation*} 
which somewhat resembles the rate distortion function \citep{ThomasCover:2006}. 
At this point, we shall leave further investigation into these directions to our future research.

\section{Conclusions and Discussion}

We have shown that there exists a relatively small group of training examples that suffices for generalization. Nevertheless, the AEP does not define this set, nor the correct coding, learning or prediction method. It just reassures us that there exists a set of the sort. How do we find the typical learning set? One option may be by
predefined or learned dictionary coding: build a training set that represents the typical set, consisting of the most common structures, in a similar manner to universal source coding based on a known dictionary \citep{ThomasCover:2006}. That said, it could be claimed that in every standard training process, when we are randomly choosing i.i.d
training data points, we are essentially building a uniform distribution. In other words, in practice, we are sampling from the empirical distribution, such that all training data points are equally probable to be chosen. Hence, we are defining a typical set, from which we expect the DNN to generalize to other data points, heuristically represented by the neural net.

\section*{Appendix A}

\textbf{Proof of Theorem 3}

We assume a training set $\Psi$, where $\{\mathbf{x}_{i}\}_{i=1}^m$ are sampled from $A^n_\epsilon(P_X)$ and labeled by some function as ground truth, $\Psi=\{\{\mathbf{y}_{i},\mathbf{x}_{i}\}_{i=1}^m : \mathbf{x}_{i} \in A^n_\epsilon(P_X)\} $. 
Assume an algorithm that minimizes the training error $\mathcal{L}_{\Psi}(\mathcal{F}_{\Psi})$ defined in (\ref{2.1}).
Assuming a discrete source $P_X(\cdot)$ that emits i.i.d sequences $x^n$ of symbols.
The estimator has access only to the observed signal $y^d$. 
For the sake of the following theoretical analysis we restrict the mapping $\mathcal{F}: \mathcal{Y}^d \rightarrow \mathcal{X}^n$ to be a surjective function. 
The goal is to prove that learning with $A^n_\epsilon(P_X)$ is equivalent to training with the entire distribution with the same generalization error.

A sequence $y^d$ of symbols is passed to the learning system. For example, the system could be an encoder-decoder, such that the encoder ``compresses'' $y^d$ into a latent space representation vector $z^k$ and sends $z^k$ into the decoder. The decoder reconstructs $x^n$ from $z^k$, as $\hat{x}^n(z^k)$.
Generally speaking, the predictor reconstruct $\hat{x}^n$ from $y^d$ and is said to be successful if $\hat{x}^n=x^n$. We consider the case where every source sequence $x^n$ is assigned a unique $z^k$. Therefore one can reconstruct $x^n$ perfectly. Note that the same latent space $z^k$ can represent different observed sequences $y^d$. This assumption is true if and only if the mapping  $\mathcal{F}: \mathcal{Y}^d \rightarrow \mathcal{X}^n$ is unique. In other words, every $x^n$ can be mapped to more than one $y^d$, but every $y^d$ can only be mapped to one $x^n$.
The goal is to prove that learning by training over the typical set $A^n_{\epsilon}(P_X)$ is sufficient. 

Denote $A=\{x^n : x^n \in  A^n_\epsilon(P_X) \}$, $B=\{x^n : x^n \notin  A^n_\epsilon(P_X) \}$.
We train the predictor with a training set $\Psi_A= \{\{y_i^d, x_i^n\}_{i=1}^m : x_i^n \in A\} $, such that the generalization error (risk) over $x^n \in A$ is at most $\varepsilon^\mathcal{F}_A$. 
Now, given some test input $y^d$, if $x^n(y^d) \in A^n_\epsilon(P_X)$ the encoder passes to the decoder the $z^k$ that represents this sequence. In the general case, the predictor deciphers $y^d$ as trained by generalization. However if $x^n(y^d) \notin A^n_\epsilon(P_X)$ we can assume some unknown output $\hat{x}^n$ (the encoder sends to the decoder some unknown $z^k$ generated by the trained learning system), with error $\varepsilon^\mathcal{F}_B$. 
The average error is upperbounded by 
\begin{equation}\label{A4.2}
\mathcal{L}(h_{\Psi})   \leq Pr[ x^n \in A^n_\epsilon(P_X) ] \varepsilon^\mathcal{F}_A + Pr[ x^n \notin A^n_\epsilon(P_X) ] \varepsilon^\mathcal{F}_B.
\end{equation}
Therefore, for sufficiently large n, 
\begin{equation}\label{A4.3}
\mathcal{L}(h_{\Psi})  \leq \varepsilon^\mathcal{F}_A.
\end{equation}

An alternative way to derive the same result is as following.
The generalization error of the trained predictor $h_{\Psi}$ is
\begin{flalign*}
\nonumber
\mathcal{L}(h_{\Psi}) & = E_{(x,y) \sim P_{X,Y}} \ \ell \big(h_{\Psi}(y),x\big) \\
\nonumber
& = \sum_{(x,y) \sim P_{X,Y}} P_{X,Y}(x,y) \ \ell\big(h_{\Psi}(y),x\big)  
\\
\nonumber
& = \sum_{x \sim P_{X}} P_X(x) \sum_{y \sim P_{Y|X}} P_{Y|X}(y|x) \ \ell\big(h_{\Psi}(y),x\big)
\\
\nonumber
& = \sum_{x \in A^n_\epsilon(P_X)} P_X(x)\sum_{y \sim P_{Y|X}} P_{Y|X}(y|x) \ \ell\big(h_{\Psi}(y),x\big) 
\\
& =
\mathcal{L}_A(h_{\Psi}) \leq \varepsilon^\mathcal{F}_A ,
\end{flalign*}
where $\mathcal{L}_A(h_{\Psi})$ denotes the generalization error over the typical set $A^n_\epsilon(P_X)$.
The third equality follows from $Pr[x^n\in A^n_\epsilon(P_X)]=1$, for sufficiently large $n$.
Throughout the proofs we sometimes omit the superscript $^n$ for simplicity. 
Note that this derivation is symmetric for $x$ and $y$, therefore it is possible to build a training set by drawing  samples $y^d\in A^d_\epsilon(P_Y)$, for sufficiently large $d$, and pairing them with the corresponding $x^n$, under the assumption that the mapping $\mathcal{F}: \mathcal{Y}^d \rightarrow \mathcal{X}^n$ is a surjective function. Alternatively, it is possible to define the jointly typical set $\mathcal{B} =\{(x^n,y^d): (x^n,y^d) \in A^{n,d}_\epsilon(P_{X,Y}), y^n \in A^d_\epsilon(P_{Y}) , x^n \in A^n_\epsilon(P_{X}) \}$, for sufficiently large $n$ and $d$, and assume the training input-output pairs are drawn from $\mathcal{B}$.
\qed 

\section*{Appendix B}

\textbf{Proof of Theorem 4}

Assume a supervised learning algorithm with a training set $\Psi$,
sampled from an unknown probability mass function $P_X$ and paired with $y$ by some target function $g(\cdot)$,
to learn a predictor $\mathcal{F}_{\Psi} : \mathcal{Y}^d \rightarrow \mathcal{X}^n $.
(Here, the subscript $\Psi$ emphasizes that the output predictor depends on $\Psi$.)
Let the learner's output sequence $\mathbf{x}\in\mathcal{X}^{n \times 1}$ be a finite sequence whose i'th entry $x_i$ takes on values in a discrete and finite alphabet $\mathcal{X}$.
We write $\mathcal{X}^n$ for the Cartesian product of the set $\mathcal{X}$ with itself $n$ times.
Recall the sequences that serve as the examples in the training set are independently
and identically distributed (i.i.d.).
The algorithm is designed to find $\mathcal{F}_{\Psi}$ that minimizes the error over an unknown 
$P_{X,Y}(\cdot)$ over $\mathcal{X}^n \times \mathcal{Y}^d $. But the true error is not
available to the learner (since $P_{X,Y}(\cdot)$ is unknown). Therefore, the learner attempts to minimize the training error $\mathcal{L}_{\Psi}(\mathcal{F}_{\Psi}) $ as defined in (\ref{2.1}).
Assume the predictor was trained successfully to yield $\mathcal{L}_{\Psi}(h_{\Psi})\leq\Delta_m << 1$.
We define the true error, or the generalization error, $\mathcal{L}(h_{\Psi})$ in (\ref{2.2}).

We are interested in finding the sample size of $m$ sequences of instances that will lead to a bounded generalization error for the specific trained predictor,
$\mathcal{L}(h_{\Psi})  \leq \Delta$.
Namely,
{\small
\begin{flalign*}
\nonumber
&\mathcal{L}(h_{\Psi}) = E_{(x,y) \sim P_{X,Y}} \ \ell \big(h_{\Psi}(y),x\big) 
= \sum_{x,y} P_{X,Y}(x,y) \ \ell\big(h_{\Psi}(y),x\big)  
\\
\nonumber
& = \sum_y P_Y(y) \ \ell\big(h_{\Psi}(y),f(y) \big)
= \sum_{y\in A^d_\epsilon(P_Y)} P_Y(y) \ \ell\big(h_{\Psi}(y),f(y)\big).
\end{flalign*}
}
The second equation follows from $P(X,Y)=P(Y)P(X|Y)$. $P(X=f(y)|Y=y)=1$ since we assumed $g(\cdot)$ is a deterministic function, and $f(\cdot)$ is a surjective function. The third equation follows from $Pr[y^d\in A^d_\epsilon]=1$, for sufficiently large $d$. If $ m \geq 2^{dH(Y)}$, and since, for small $\epsilon$, $|A_\epsilon^d(P_Y)|=2^{dH(Y)}$, and $P_Y(y^d)=2^{-dH(Y)}$, then clearly $\mathcal{L}(h_{\Psi})=\mathcal{L}_{\Psi}(h_{\Psi}) \leq \Delta_m$.
Otherwise, $m < 2^{dH(Y)}$ and
\begin{equation*}
\nonumber
\mathcal{L}(h_{\Psi})
\leq (2^{dH(Y)}-m)2^{-dH(Y)}+m \Delta_m 2^{-dH(Y)}\leq \Delta,
\end{equation*}
where we have used the fact that $\ell(x,\hat{x})\leq 1$. 
We now have
\begin{equation}
1-m2^{-dH(Y)}(1-\Delta_m) \leq \Delta,
\end{equation}
Therefore,
\begin{equation}\label{A.8}
m \geq 2^{dH(Y)}\frac{1-\Delta}{1-\Delta_m}.
\end{equation}
Equivalently, 
\begin{equation}
\Delta \geq 1-\frac{m}{2^{dH(Y)}}(1-\Delta_m).
\end{equation}
The ratio between the sample size and the input data typical set determines the upper bound on the generalization error. 
If one wishes, since $\Delta_m<<1$, (\ref{A.8}) can be written as 
\begin{equation}
m \geq 2^{dH(Y)}\frac{1-\Delta}{1-\Delta_m}
\approx 2^{dH(Y)}(1-\Delta)(1+\Delta_m),
\end{equation}
where the approximation is derived using a first order Taylor's expansion. 
Hence, approximately we have
\begin{equation}
m \geq 2^{dH(Y)}(1-\Delta).
\end{equation}
\qed

\section*{Appendix C}

\textbf{Proof of Theorem 5}

In this case, we assume $y=g(x)+w$, such that each possible $y$ input induces a conditional probability mass function over the possible $x$ outputs. When the same $y$ could originate in two different output $x$'s, then the $x$'s outputs are indistinguishable, and therefore could not be efficiently learned. Since we assumed a surjective mapping $\mathcal{F}: \mathcal{Y}^n \rightarrow \mathcal{X}^n$, for a distinguishable subset of input sequences, there exists only one $x^n$ that could have caused a particular $y^n$ with high probability. We can therefore reconstruct the
sequence at the output with a negligible probability of error, by
mapping the observations into the proper ``widely spaced'' hidden sequences.
We can define the conditional entropy $H(Y|X)$ assuming they are ergodic and have a stationary coupling \citep{Gray:2011}.
Defining their mutual information $I(X;Y)=H(Y)-H(Y|X)$, their jointly typical
set follows similar properties \citep{ThomasCover:2006}.
Define $B=\{(x^n,y^n): (x^n,y^n) \in A^n_\epsilon(P_{X,Y}), y^n \in A^n_\epsilon(P_{Y}) , x^n \in A^n_\epsilon(P_{X}) \}$,and $A^n_\epsilon(P_{X,Y}|x^n)= \{y^n : (x^n,y^n) \in A^n_\epsilon(P_{X,Y})\}$.
Observe that $A^n_\epsilon(P_{X,Y}|x^n)=\emptyset$ if $x^n \notin A^n_\epsilon(P_{X})$  \citep{Kramer:2008}.
For $n$ sufficiently large and small $\epsilon$,
\begin{equation}\label{C3.1}
\mathrm{Pr}[Y^n \in A^n_\epsilon(P_{X,Y}|x^n)|X^n=x^n] = 1.
\end{equation}
\begin{equation}\label{C3.2}
\mathrm{Pr}[(x^n,y^n) \in B] = 2^{-nI(X;Y)}.
\end{equation}
\begin{equation}\label{C3.3}
|B| = 2^{nI(X;Y)}.
\end{equation}
Roughly speaking, if $y^n$ and $x^n$ are jointly typical, then we can resolve an input $y^n$ as $x^n$. There are approximately $2^{nH(Y|X)}$ equally probable $y^n$ sequences, for each typical output sequence $x^n$. We assume
that no two $x^n$ sequences correspond to the same $y^n$ output sequence,
otherwise, the learner will not be able to decide which $x^n$ sequence it originated from.
There are approximately $2^{nH(Y)}$ possible typical $y^n$ sequences. This set
is split into sets of size $2^{nH(Y|X)}$,
associated with different $x^n$ sequences. Therefore, the total number of distinguishable sets is less than or equal
to {\small $2^{n(H(Y)-H(Y|X))}=2^{nI(X;Y)}$}.
Hence, we can have at most $2^{nI (X;Y)}$ disjoint sequences of length $n$.

Therefore, 
{\small
\begin{flalign*}
\mathcal{L}(h_{\Psi}) 
& = E_{(x,y) \sim P_{X,Y}} \ \ell \big(h_{\Psi}(y),x\big) 
= \sum_{x,y} P_{X,Y}(x,y) \ \ell\big(h_{\Psi}(y),x\big)  \\
& = \sum_{x,y\in B} P_{X,Y}(x,y) \ \ell\big(h_{\Psi}(y),x\big).
\end{flalign*}
}The last equation follows from (\ref{C3.1}).
If $m \geq 2^{nI(X;Y)}$, and since $|B|=2^{nI(X;Y)}$, and $\mathrm{Pr}[(x^n,y^n) \in B]=2^{-nI(X;Y)}$, then clearly $\mathcal{L}(h_{\Psi})=\mathcal{L}_{\Psi}(h_{\Psi}) \leq \Delta_m$.
Otherwise, when $m < 2^{nI(X;Y)}$, we have
\begin{equation*}
\mathcal{L}(h_{\Psi})
\leq (2^{nI(X;Y)}-m)2^{-nI(X;Y)}+ m \Delta_m 2^{-nI(X;Y)} \leq \Delta,
\end{equation*}
assuming $\ell(x,\hat{x})\leq 1$. 
We now have
\begin{equation}
m2^{-nI(X;Y)}(1-\Delta_m) \geq 1- \Delta,
\end{equation}
Equivalently 
\begin{equation}
\Delta \geq 1- \frac{m}{2^{nI(X;Y)}}(1-\Delta_m).
\end{equation}
Assuming $1-\Delta_m \approx 1$, 
\begin{equation}
m \geq 2^{nI(X;Y)}(1-\Delta).
\end{equation}
\qed

\section*{Acknowledgments}
The author thanks Uzi Pereg (Technion - Israel Institute of Technology) for helpful discussions. The author thanks the associate editor and the anonymous reviewers for their constructive comments and valuable
suggestions. This work was supported in part by the Zuckerman STEM Leadership
Program.
\bibliography{dpereg_iclr2023}

\end{document}